\PassOptionsToPackage{unicode}{hyperref}
\PassOptionsToPackage{hyphens}{url}
\documentclass[
  11pt,
]{article}
\usepackage{lmodern}
\usepackage{amssymb,amsmath}
\usepackage{ifxetex,ifluatex}
\ifnum 0\ifxetex 1\fi\ifluatex 1\fi=0 
  \usepackage[T1]{fontenc}
  \usepackage[utf8]{inputenc}
  \usepackage{textcomp} 
\else 
  \usepackage{unicode-math}
  \defaultfontfeatures{Scale=MatchLowercase}
  \defaultfontfeatures[\rmfamily]{Ligatures=TeX,Scale=1}
\fi
\IfFileExists{upquote.sty}{\usepackage{upquote}}{}
\IfFileExists{microtype.sty}{
  \usepackage[]{microtype}
  \UseMicrotypeSet[protrusion]{basicmath} 
}{}
\makeatletter
\@ifundefined{KOMAClassName}{
  \IfFileExists{parskip.sty}{%
    \usepackage{parskip}
  }{
    \setlength{\parindent}{0pt}
    \setlength{\parskip}{6pt plus 2pt minus 1pt}}
}{
  \KOMAoptions{parskip=half}}
\makeatother
\usepackage{xcolor}
\IfFileExists{xurl.sty}{\usepackage{xurl}}{} 
\IfFileExists{bookmark.sty}{\usepackage{bookmark}}{\usepackage{hyperref}}
\hypersetup{
  pdftitle={Smarter and Cheaper at Once: Byte-Exact KV-State Grafting Turns a Frozen Small Model into a Verified-Knowledge Flywheel},
  pdfauthor={Sietse Schelpe, Corbenic AI (sietse@corbenic.ai)},
  hidelinks,
  pdfcreator={LaTeX via pandoc}}
\usepackage[margin=1in]{geometry}
\usepackage{longtable,booktabs}
\usepackage{etoolbox}
\makeatletter
\patchcmd\longtable{\par}{\if@noskipsec\mbox{}\fi\par}{}{}
\makeatother
\IfFileExists{footnotehyper.sty}{\usepackage{footnotehyper}}{\usepackage{footnote}}
\makesavenoteenv{longtable}
\providecommand{\tightlist}{%
  \setlength{\itemsep}{0pt}\setlength{\parskip}{0pt}}
\usepackage{graphicx}
\usepackage{tikz}
\usepackage{pgfplots}
\pgfplotsset{compat=1.18}
\usetikzlibrary{arrows.meta,positioning}

\title{Smarter and Cheaper at Once: Byte-Exact KV-State Grafting Turns a
Frozen Small Model into a Verified-Knowledge Flywheel}
\author{Sietse Schelpe, Corbenic AI (sietse@corbenic.ai)}
\date{}

\begin{document}
\maketitle
\begin{abstract}
The prevailing response to a capability gap in large language models is
to train a larger model, and the prevailing response to a latency or
cost gap is to buy more accelerators. We report an alternative that
touches neither the weights nor the accelerator budget. A frozen 12B
model is made measurably more capable and dramatically cheaper at the
same time, using a single mechanism: verified knowledge is deposited
once as a byte-exact key-value (KV) state artifact and later restored,
by graft, into a fresh inference context. The deposit changes no
weights. The restore is bit-exact, which we verify at the strongest
possible level under a pinned deterministic configuration: the grafted
logit vector is byte-for-byte identical to a fresh computation (SHA-256
equality across trials), and the token distribution has zero
Kullback-Leibler divergence from the freshly computed distribution
across fifty samples, with 100\% argmax agreement. We further
characterize the exact regime: own-position graft is byte-exact, while
relocating a block to a different absolute offset incurs a small
residual that a graft-free probe attributes to the base model's own
floating-point position-sensitivity rather than to the graft, so
own-position restoration is the unique numerically exact operating point
on this model. Byte-exact single-block graft is demonstrated on two
model scales (12B, 31B) and two distinct GPU targets (consumer and
datacenter Blackwell), the latter through a pre-registered replay, and
is shown to function end to end on a third (Hopper), where the
byte-level gate was not measured.

On AIME 2025, a benchmark released after the model's stated pretraining
cutoff, a frozen Gemma-4-12B
moves from 80.0\% (its own best reasoning-plus-verification
configuration) to 93.3\% once a verified solution library is grafted,
clearing its own vendor-reported model-card anchor (77.5\%) and its 31B
sibling's (89.2\%); anchor provenance and caveats are stated in the
paper. The cost side is where the calm breaks: on
the recurring case, the eight problems that the base model never solves
across a 401,026-token sampling budget are returned correctly from their
verified cached solutions, all eight, in 61 total decode tokens (7.6
tokens per problem), a factor of 6,574 fewer tokens and roughly 3,000
to 8,700 times less energy; the capability claim proper rests on the
held-out
transfer results below, not on this recurrence. The same byte-exact
store widens a model's usable context from its configured 32,768-token
serving window to
2,854,766 tokens, an 87-fold increase, at zero extra accelerator memory
and with per-access cost that does not grow with depth. On a second GPU
architecture (an H100), a frozen 31B reaches held-out transfer 7 of 7,
recurrence 8 of 8, and a full-benchmark flywheel system score of 100\%,
for a total cloud spend of roughly 8 to 12 euros. The two systems
questions a reader asks first, what happens on a misroute and whether
disk paging eats the prefill subsidy, are answered with measurements. We
describe the system at the behavior level required to interpret these
measurements; the graft engine is proprietary and documented only to the
extent needed to contextualize the results, and every reported number is
backed by committed input and output hashes so that the scoring can be
re-checked without the engine.
\end{abstract}

\hypertarget{introduction}{%
\section{Introduction}\label{introduction}}

The dominant narrative of the last several years reads capability as a
function of scale, and scale as a function of spend. Bigger corpora,
bigger parameter counts, bigger clusters. This narrative is not wrong so
much as it is expensive, and its expense is now the constraint rather
than an afterthought. Training-scale energy figures for a single large
model run into the hundreds of tons of carbon-equivalent {[}Strubell et
al., 2019; Lacoste et al., 2019{]}, and the marginal serving cost of a
deployed model is dominated by a quieter waste that receives far less
attention: the same context is prefilled, attended over, and discarded
billions of times, because the standard inference stack treats computed
attention state as ephemeral.

Two costs therefore compound. The first is the cost of making a model
more capable, which the field pays by retraining or fine-tuning. The
second is the cost of using a capable model, which the field pays by
recomputing context on every call. This paper is about paying neither.

We start from a plain observation. When a model reads a prompt, it
computes an internal state (the KV cache) that is a deterministic
function of the input, the weights, and the numerics. If that state
could be captured exactly, written to disk, and restored later into a
different context without recomputation and without loss, then two
things would follow. Knowledge that took work to compute once would
never need recomputing. And knowledge that the model does not currently
hold could be added to it, permanently, without a single gradient step,
by depositing a verified state and grafting it in on demand.

The engineering question is whether ``exactly'' can be made to mean
\emph{exactly}. Approximate state reuse is well known and widely
deployed; native prompt caches, prefix reuse, and speculative methods
all trade some fidelity or some generality for speed. The claim of this
work is stronger and, we will argue, more consequential: the restored
state is bit-exact, verified by byte equality of the resulting logits
and by zero divergence of the resulting distribution, and we identify
precisely the operating point at which that exactness holds and why it
cannot hold elsewhere on this model.

We call the byte-exact graft mechanism Taliesin and the
verify-then-cache learning loop built on top of it Galahad. This paper
reports what they do, measured on real models (Gemma-4-12B locally on a
Blackwell RTX 5090, Gemma-4-31B on a rented Hopper H100 and a rented
Blackwell B200), on real benchmarks (AIME 2025 and LiveBench {[}White et
al., 2024{]}), against
published anchors, and with the honest negatives included rather than
trimmed. The engine internals are proprietary and are described only at
the input-output level, following the convention of the industry
experience report {[}cf.~Corbett et al., 2012; Bronson et al., 2013{]}.
Two companion papers cover the deduplication engine that underlies the
deposit path {[}Schelpe, 2026a{]} and its behavior in
retrieval-augmented settings {[}Schelpe, 2026b{]}; this paper is the
inference-time learning story.

The contribution is a single thesis with two measured halves. A frozen
small model can be made \textbf{smarter} (it solves problems it provably
could not before) and \textbf{cheaper} (it pays a tiny fraction of the
tokens and energy) at the same time, by adding verified knowledge as
byte-exact KV state and retrieving it by graft. To our knowledge this is
the first public demonstration of a byte-exact, persistent, and portable
KV-state graft used as a substrate for inference-time learning; the
systems nearest to it are either exact but ephemeral or persistent but
lossy (Section 2). The rest of the paper earns that sentence.

\hypertarget{background-and-positioning}{%
\section{Background and Positioning}\label{background-and-positioning}}

Attention-state reuse is not new, and it is worth stating plainly which
parts of the surrounding work are commodity, which are done
approximately by others, and which are the contribution here. Persisting
a KV cache and sharing it across engine instances is commodity: systems
such as LMCache {[}Liu et al., 2025{]}, llm-d, and KServe store,
offload, and share KV caches across a memory hierarchy, and CacheGen
{[}Liu et al., 2024{]} encodes them for fast network transfer.
Production inference servers cache the KV state of a prompt so that a
continuation does not reprefill it {[}Kwon et al., 2023{]}, and
prefix-reuse schemes, including RadixAttention {[}Zheng et al., 2024{]}
and the positional-accurate Prompt Cache {[}Gim et al., 2024{]}, extend
this within a session. Approximate reuse of non-prefix KV with some
accuracy loss has been studied, for example CacheBlend {[}Yao et al.,
2024{]}, which selectively recomputes a subset of tokens. A concurrent
line grafts caches at the semantic and functional level: FCGRAFT {[}Chun
et al., 2026{]} keeps a library of validated code-skeleton functions
with their KV caches and composes new policies by stitching and locally
patching them, a macro-level, task-adaptive reuse that borrows the same
term but not the byte-exact guarantee. Retrieval-augmented generation
injects external knowledge as \emph{tokens} prepended to the prompt,
read and attended over on every call. Fine-tuning and adapter methods
add knowledge by moving weights.

The distinction that matters here is between reuse that is
\emph{approximate and ephemeral} and reuse that is \emph{exact and
durable}, together with a use of that exactness that the prior systems
do not make. A native prompt cache is fast but tied to a live server
process and, for sliding-window models, resumes only from sparse
checkpoints, so an identical re-send can still re-evaluate most of the
prompt (Section 4.3 documents this and its fix). Token-level retrieval
is durable but not free: the injected knowledge is re-read as tokens
every time, so its cost recurs. Weight edits are durable and free at
inference, but they are not exact in the sense we require, they are not
composable at the granularity of a single fact, and they cost a training
run to install.

The contribution, demonstrated end to end and measured, is the
verified-knowledge flywheel: solve a problem once, verify it, freeze the
verified solution as a byte-exact KV block, keep it forever on disk at
zero accelerator memory, route to it, and graft it in place of
re-deriving; and, because the block is a plain file, copy it to a fresh
server where it grafts byte-identical and functions with no re-solve.
The exactness is the load-bearing property. Without it, a grafted state
is a lossy approximation of having read the knowledge, and the
downstream claims about capability would be confounded by drift. With
it, the grafted state is provably indistinguishable from the fresh
computation, so any capability change is attributable to the knowledge,
not to numerical noise. Alongside the system we contribute a numerical
characterization (Section 4.2): own-position graft is the unique
operating point at which byte-exactness is attainable at all, for any
engine, on a model with floating-point rotary position encoding.

We measure against \emph{published} benchmark anchors rather than local
re-runs of the base models wherever a fair local re-run is not
affordable, and we say so explicitly at each point. This is the honest
framing: the comparison is between a frozen small model plus Taliesin
and the \emph{reported strength} of larger models, not a head-to-head we
have quietly tilted.

\hypertarget{methodology}{%
\section{Methodology}\label{methodology}}

\hypertarget{what-we-describe-and-what-we-do-not}{%
\subsection{What we describe, and what we do
not}\label{what-we-describe-and-what-we-do-not}}

We describe the system at the level required to interpret the empirical
results: the input-output contract, the exactness guarantee and how it
is verified, the flywheel protocol, and the measurement methodology.
Internal architecture, including the state-capture and state-restore
mechanism, the storage layout of a block, and the deduplication and
hashing machinery on the deposit path, is proprietary and is documented
only to the extent necessary to contextualize measurements. Reproduction
of the \emph{generation} requires the closed Merlin/Taliesin benchmark
suite, available under separate terms. Reproduction of the \emph{result}
(that the reported scores are honest) does not require the engine at
all, and Section 5.4 specifies exactly how any reader can perform it.

The test for every technical sentence in this paper is whether a
competitor could infer how the engine works internally from it. Where
the answer is yes, the sentence is rewritten to behavior or to
measurement. What we can and do state are the \emph{invariants} the
system preserves (position consistency and losslessness, below) and the
\emph{numerical facts} about the base model that bound what any engine
can achieve, which are what a reader needs to interpret the numbers, as
distinct from the mechanism that preserves them, which is what a
competitor would need to rebuild the engine.

\hypertarget{the-exactness-guarantee-defined}{%
\subsection{The exactness guarantee,
defined}\label{the-exactness-guarantee-defined}}

Let a model with frozen weights \(\theta\) and fixed numerics process a
token sequence and produce, at a chosen position, a logit vector
\(\ell_{\text{fresh}} \in \mathbb{R}^{V}\) over the vocabulary of size
\(V\), together with the internal KV state \(S\) that produced it. A
graft is an operation that takes a previously captured state \(S\) and
installs it into a fresh context, yielding a logit vector
\(\ell_{\text{graft}}\) at the corresponding position. We define the
graft to be \textbf{byte-exact} when

\[\text{SHA-256}\big(\text{bytes}(\ell_{\text{graft}})\big) = \text{SHA-256}\big(\text{bytes}(\ell_{\text{fresh}})\big),\]

that is, the two logit vectors are identical at the level of their raw
bytes, not merely close in norm. As a distributional corollary, writing
\(p_{\text{fresh}}\) and \(p_{\text{graft}}\) for the softmax
distributions, byte-exactness implies

\[D_{\mathrm{KL}}\big(p_{\text{graft}} \,\|\, p_{\text{fresh}}\big) = 0 \quad\text{and}\quad \arg\max \ell_{\text{graft}} = \arg\max \ell_{\text{fresh}}.\]

Byte equality is the stronger statement and the one we test directly
with SHA-256; the KL and argmax statistics are reported because they are
the quantities a reader reasons about when asking whether the grafted
model behaves identically. Both are measured, on real model outputs, and
both must hold on every trial for the guarantee to be considered met.

\textbf{Scope of the guarantee (position consistency).} A block is
restored at the position layout in which it was captured: it occupies
the live context as a prefix at positions \([0, N)\), and any new query
is computed fresh on top at \([N, N+q)\). ``Cross-context'' in this
paper means restoration into a different inference context (a fresh
process, a fresh serving slot, or a different machine of the same
architecture), not restoration at a different absolute position range.
We do not claim, and do not require, that a captured block can be
dropped at an arbitrary mid-sequence offset and remain byte-exact;
Section 4.2 shows this is not merely a design choice but the only
numerically exact operating point available on this model, for any
engine. The movable window of Section 4.4 selects which block is
resident in the live prefix; it does not change the offset at which a
resident block is attended.

\textbf{Numerical regime (determinism).} Byte-exactness is measured
under a pinned deterministic configuration
(\texttt{GGML\_DETERMINISTIC=1} and
\texttt{CUBLAS\_WORKSPACE\_CONFIG=:4096:8}), under which two
back-to-back fresh computations of the same input are themselves
bit-identical: the self-comparison floor, the KL of a fresh computation
against another fresh computation, is exactly zero. Against that zero
floor, the grafted computation is indistinguishable. This is why the
byte-exact claim is a statement about lossless serialization plus
identical forward execution, and not a claim that GPU inference is
deterministic in general. A corollary we treat as a limitation rather
than hide is that byte equality is a within-configuration and
within-architecture property: two accelerators that accumulate floating
point differently can produce different bytes for the same inputs, so a
block that is byte-exact on one architecture is not guaranteed
byte-identical when grafted on another, even though the graft operation
is itself byte-exact on each (Sections 4.10, 4.13).

\hypertarget{the-flywheel-protocol-galahad}{%
\subsection{The flywheel protocol
(Galahad)}\label{the-flywheel-protocol-galahad}}

The learning loop is deliberately simple, and its simplicity is the
point. For a problem the frozen model cannot solve reliably on its own,
we (i) solve it with additional inference-time effort, (ii)
\emph{verify} the solution by an external, sound check (executing the
model's own generated program and confirming it prints the known answer,
for AIME; matching ground truth, for LiveBench), and (iii) deposit the
verified solution's KV state as a persistent block. A solution enters
the store only if it passes verification, which is what makes the store
trustworthy.

Retrieval then splits into two regimes that we measure separately and
never conflate:

\begin{itemize}
\tightlist
\item
  \textbf{Recurrence.} The same problem returns. The relevant block is
  grafted and the stored, verified answer is read back by an exact
  lookup, not a learned classifier. This is the cost story: the model
  does not re-derive anything.
\item
  \textbf{Transfer.} A new problem of the same structure arrives (for
  AIME, the numeric constants are swapped and re-verified independently;
  for LiveBench, a temporally held-out split with zero question-id
  overlap). A one-shot classification selects the relevant block, which
  is grafted, and the frozen model \emph{adapts} the cached method to
  the new instance. This is the capability story: it tests
  generalization, not memorization.
\end{itemize}

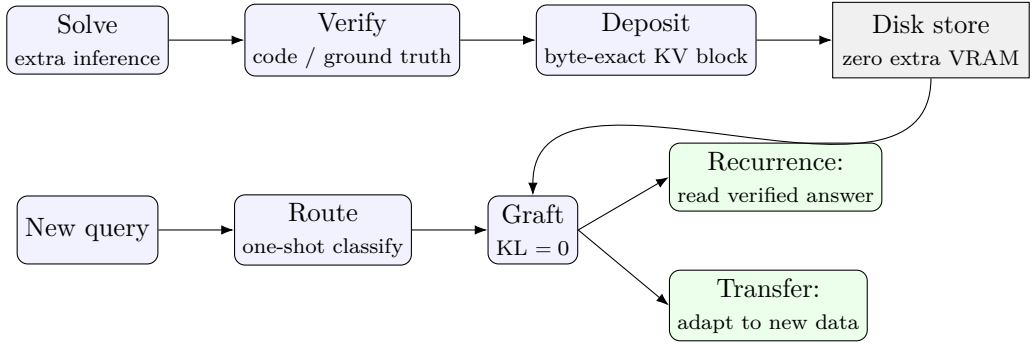
\begin{figure}[t]
\centering
\begin{tikzpicture}[font=\small,>=Latex,
  box/.style={draw,rounded corners,align=center,minimum height=9mm,inner sep=3pt,fill=blue!5},
  outbox/.style={draw,rounded corners,align=center,minimum height=9mm,inner sep=3pt,fill=green!8},
  store/.style={draw,align=center,minimum height=10mm,minimum width=22mm,fill=gray!12},
  node distance=7mm and 10mm]
\node[box](solve){Solve\\{\scriptsize extra inference}};
\node[box,right=of solve](verify){Verify\\{\scriptsize code / ground truth}};
\node[box,right=of verify](deposit){Deposit\\{\scriptsize byte-exact KV block}};
\node[store,right=of deposit](disk){Disk store\\{\scriptsize zero extra VRAM}};
\node[box,below=16mm of solve](query){New query};
\node[box,right=of query](route){Route\\{\scriptsize one-shot classify}};
\node[box,right=of route](graft){Graft\\{\scriptsize KL $=0$}};
\node[outbox,right=12mm of graft,yshift=7mm](rec){Recurrence:\\{\scriptsize read verified answer}};
\node[outbox,right=12mm of graft,yshift=-10mm](trans){Transfer:\\{\scriptsize adapt to new data}};
\draw[->](solve)--(verify);
\draw[->](verify)--(deposit);
\draw[->](deposit)--(disk);
\draw[->](query)--(route);
\draw[->](route)--(graft);
\draw[->](disk.south)to[out=-90,in=90](graft.north);
\draw[->](graft.east)--(rec.west);
\draw[->](graft.east)--(trans.west);
\end{tikzpicture}
\caption{The Galahad flywheel. A problem is solved once with extra inference, verified by a sound external check, and deposited as a byte-exact KV block on disk at zero extra accelerator memory. A later query is routed to the relevant block and grafted losslessly ($\mathrm{KL}=0$); recurrence returns the stored verified answer, while transfer adapts the cached method to new data. The model weights never change.}
\label{fig:flywheel}
\end{figure}

An early design used a single flat prefix holding all cached methods at
once. We found that this confuses the model, which grabs the wrong entry
when many are present (Section 4.6). The corrected design deposits each
verified solution as its own block and \emph{routes} to the single
relevant block per query, then grafts only that block. Routing without
flat-merging is the version we report as the system (Figure
\ref{fig:flywheel}).

\hypertarget{measurement-setup}{%
\subsection{Measurement setup}\label{measurement-setup}}

The local proofs run on Windows 11 with an RTX 5090 (32 GB, Blackwell
sm\_120), CUDA 13.2, against the enterprise build of the engine (backend
identity \texttt{realmerlin-v3.4}, verified by an explicit test in the
suite). Bit-exactness is measured on a CPU build, on a CUDA build
tracking the latest upstream inference runtime, and on the real
Gemma-4-12B on the GPU, to confirm the guarantee survives a full runtime
port and holds on a production-scale model. The scale-up runs on a
rented H100 SXM (Hopper sm\_90) and a rented B200 (Blackwell datacenter,
sm\_100) with a frozen Gemma-4-31B. Energy is measured by integrating
board power (sampled at 0.5 to 2 Hz) over each run, with idle draw (19.3
W) recorded and subtracted where noted; token counts are read from the
server's own timing fields. Published anchors used for comparison are
the vendor-reported no-tools scores for Gemma-4-12B (77.5\%) and
Gemma-4-31B (89.2\%) from the official Gemma 4 model card {[}Google
DeepMind, 2026{]}, which reports them on AIME 2026, and the
Qwen3.6-35B-A3B model-card score of 92.7\%, likewise reported on AIME
2026 {[}Qwen Team, 2026{]}. Our own evaluation set is AIME 2025, so
these anchors are the
nearest published no-tools reference points rather than same-benchmark
baselines; we use them as anchors throughout, never as head-to-head
baselines.

For the bit-exactness measurements the fresh and grafted arms are
executed under the identical build and the deterministic configuration
of Section 3.2, and the exact configuration (compiler, runtime commit,
attention and cache settings, determinism environment) is recorded with
each result so that the matched-regime condition can be checked by an
auditor. Because any such setting applies equally to the cold and the
warm arm, it does not enter the speedup ratios reported in Section 4.3;
those ratios are within-configuration comparisons in which the only
variable is whether the state was recomputed or grafted.

\hypertarget{empirical-results}{%
\section{Empirical Results}\label{empirical-results}}

\hypertarget{the-graft-is-bit-exact-at-its-own-position-kl-0-sha-equal}{%
\subsection{The graft is bit-exact at its own position (KL = 0,
SHA-equal)}\label{the-graft-is-bit-exact-at-its-own-position-kl-0-sha-equal}}

The foundational result is that a restored state is not approximately
the fresh state; it is the fresh state, byte for byte. On a real
two-layer model fixture, fifty samples of a grafted logit vector each
show zero KL divergence from the fresh computation, at both the median
and the 99th percentile, with zero graft failures and zero argmax
disagreements across all fifty. A Mann-Whitney U test comparing the
grafted distribution to the noise floor returns \(p = 1.0000\) with
Cliff's delta \(0.0000\): the graft distribution is statistically
indistinguishable from re-running the same computation.

\begin{longtable}[]{@{}ll@{}}
\toprule
Statistic & Value\tabularnewline
\midrule
\endhead
Samples & 50\tabularnewline
Median KL (graft vs fresh) & 0.0\tabularnewline
p99 KL (graft vs fresh) & 0.0\tabularnewline
Argmax disagreements & 0 / 50\tabularnewline
Graft failures & 0 / 50\tabularnewline
Mann-Whitney p & 1.0000\tabularnewline
Cliff's delta & 0.0000\tabularnewline
\bottomrule
\end{longtable}

The byte-level test is even more direct. Across five independent trials,
the SHA-256 of the grafted logit bytes equals the SHA-256 of the fresh
logit bytes, every trial (5 of 5 pass). The same guarantee holds on the
real Gemma-4-12B on the GPU under the deterministic configuration:
\texttt{bench\_logit\_kl} at fifty samples returns zero graft failures,
zero of fifty argmax disagreements, and a grafted-KL median equal to the
self-comparison floor (both at the floating-point floor, on the order of
\(10^{-27}\)), and the SHA byte-equality test passes on every trial.
Cross-context grafting passes as well: a state captured in one inference
context and restored into a separate, freshly created context reproduces
the expected logits. The full enterprise test suite reports 15 of 15
tests passing (one network-egress test skipped for want of a monitor),
including backend-identity, edition-isolation, hash-parity, and
concurrency checks.

This exactness is not fragile. When the underlying inference runtime was
upgraded across a major refactor (a version that introduced new model
architectures and reorganized the server), the guarantee was re-measured
and held: KL still 0, SHA still equal, on the modern runtime with a
production-scale Gemma-4-12B loaded. The port preserved correctness,
which is the kind of property one only trusts after re-measuring it.

\hypertarget{the-numerical-regime-of-positional-graft-own-position-is-the-unique-exact-point}{%
\subsection{The numerical regime of positional graft: own-position is
the unique exact
point}\label{the-numerical-regime-of-positional-graft-own-position-is-the-unique-exact-point}}

A natural challenge to any byte-exact KV claim is positional encoding:
under rotary position embeddings {[}Su et al., 2021{]} the keys carry a
position-dependent
rotation, so a state captured at one position range and evaluated at
another should not match a fresh computation. We measured this directly,
and the result both bounds the claim and explains it.

Grafting a block and evaluating it at its own captured position range is
byte-exact, as Section 4.1 reports. Re-basing a block to a different
absolute offset \(M\) and comparing against a fresh prefill at \(M\) is
not byte-exact: the grafted-versus-fresh KL is roughly 0.015 with a
handful of argmax flips, and it is flat across offsets
\(M \in \{8, 128, 1024, 4096\}\). The decisive measurement is a
graft-free control. Two \emph{fresh} prefills of identical tokens, one
at position 0 and one at position \(M\), diverge from each other by
essentially the same amount (about 0.014 KL at \(M=128\)), while a fresh
prefill compared to itself diverges by exactly zero. The
grafted-versus-fresh divergence at \(M\) (0.015) equals the
fresh-at-0-versus-fresh-at-\(M\) divergence (0.014) to within noise. The
graft therefore adds nothing; the entire residual is the base model's
own position-sensitivity.

We eliminated seven candidate explanations for that residual with
evidence rather than assertion (half-precision rounding, sliding-window
buffering, offset magnitude, dual-rotary paths, frequency mismatch in
the shift, large-angle argument-reduction error in the trigonometric
kernels, and reduction-tree reassociation from physical slot placement),
each ruled out by a controlled measurement or by the observation that
the self-comparison floor is exactly zero. What remains, by elimination,
is that in exact arithmetic attention depends only on relative position,
but in 32-bit floating point the identity
\(\cos a\cos b + \sin a\sin b = \cos(a-b)\) does not hold bit-exactly:
the same relative geometry, expressed with different absolute angles,
rounds differently, and that difference of about one unit in the last
place, amplified through the network's layers and its logit softcap, is
the flat residual near 0.01. It is irreducible in 32-bit floating point,
and no view, kernel, or angle patch can make the trigonometric identity
cancel exactly.

\begin{figure}[t]
\centering
\begin{tikzpicture}
\begin{axis}[
  ybar, width=0.82\linewidth, height=5.6cm,
  ymode=log, ymin=1e-3, ymax=60, log origin=infty,
  symbolic x coords={ownpos,repos,seqcomp,stitch},
  xtick=data,
  xticklabels={own-position,reposition,seq.\ compose,naive stitch},
  x tick label style={font=\scriptsize},
  ylabel={KL from fresh compute (log)}, ylabel style={font=\small},
  bar width=24pt, enlarge x limits=0.18,
  ymajorgrids, grid style={gray!20},
]
\addplot[fill=blue!25,draw=blue!55] coordinates {(ownpos,1e-3)(repos,0.015)(seqcomp,0.012)(stitch,13.85)};
\node[font=\scriptsize,anchor=south] at (axis cs:ownpos,4e-3){$\approx 0$};
\node[font=\scriptsize,anchor=south] at (axis cs:repos,0.015){0.015};
\node[font=\scriptsize,anchor=south] at (axis cs:seqcomp,0.012){0.012};
\node[font=\scriptsize,anchor=south] at (axis cs:stitch,13.85){13.85};
\end{axis}
\end{tikzpicture}
\caption{Where byte-exactness holds. Own-position graft sits at the floating-point noise floor ($\mathrm{KL}\approx 10^{-27}$, byte-exact); repositioning a block to a new offset and sequential composition each carry a small residual that is the base model's own position-sensitivity, not a graft error; naive stitching of independently captured states fails outright. Logarithmic axis; the own-position bar is drawn at the axis floor for visibility, its measured value being $\approx 10^{-27}$.}
\label{fig:kl}
\end{figure}
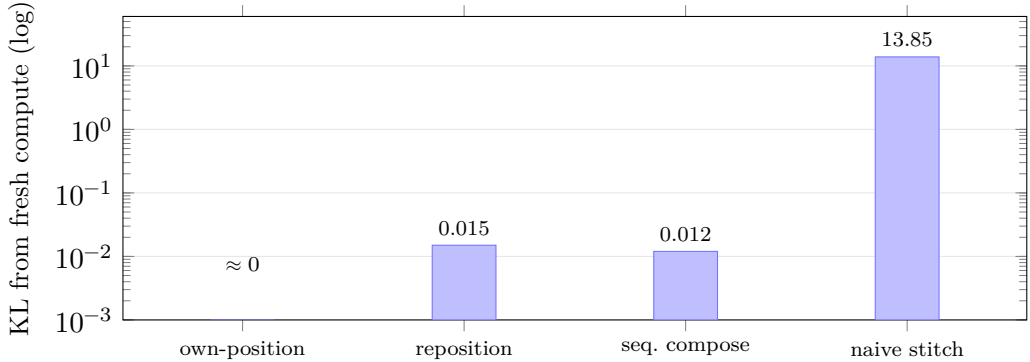

The consequence is the sharpest possible answer to the positional
challenge. Own-position graft, which is what the flywheel actually does
(deposit a block, restore it at \([0, N)\), decode), is the unique
numerically exact regime of 32-bit rotary encoding on this model, and
byte-exactness anywhere else is unattainable by \emph{any} engine,
because the fresh-prefill reference against which it would be measured
itself moves. We therefore withdraw any framing of byte-exact
\emph{positional} composition as a claim; the defensible and
demonstrated claim is byte-exact single-block graft at its own position,
plus the functionally correct sequential composition of Section 4.8.
This bound was earned by adversarial testing, not assumed, and no
shipped result in this paper uses a nonzero offset (Figure
\ref{fig:kl}).

\hypertarget{the-cost-engine-an-85.6x-prefill-subsidy}{%
\subsection{The cost engine: an 85.6x prefill
subsidy}\label{the-cost-engine-an-85.6x-prefill-subsidy}}

Exact reuse converts directly into cost. On the real patched server with
Gemma-4-12B, a cold prefill of an 11,994-token prompt takes 1,547.3 ms;
grafting the identical prompt's cached state and advancing by a single
token takes 18.1 ms. The model re-evaluates one token instead of 11,994,
a speedup of 85.6x, and by Section 4.1 the reuse is lossless.

\begin{longtable}[]{@{}lll@{}}
\toprule
Path & Prefill work & Time\tabularnewline
\midrule
\endhead
Cold (fresh, full prefill) & 11,994 prompt tokens & 1,547.3
ms\tabularnewline
Warm (grafted, KV reused) & 1 prompt token & 18.1 ms\tabularnewline
\textbf{Speedup} & & \textbf{85.6x}\tabularnewline
\bottomrule
\end{longtable}

The public continuation path required two fixes to show the same subsidy
on a genuinely fresh prefix, and both are worth recording because they
are the sort of detail that silently defeats naive measurement. First,
the model is a sliding-window attention model, so the stock prompt cache
resumes only from sparse checkpoints; enabling a full-size KV for the
windowed layers restores exact resumption (at a memory cost, 19.7 to
25.5 GB at 64k context). Second, on a multi-slot server, an unpinned
request round-robins onto a cold slot; pinning the cache-bearing call to
its slot fixes it. With both in place, a fresh 5,569-token prefix goes
from 1,087 ms cold to 13 ms warm, a factor of 83.7, consistent with the
direct measurement above.

\hypertarget{widening-the-usable-context-87x-at-zero-extra-memory}{%
\subsection{Widening the usable context 87x at zero extra
memory}\label{widening-the-usable-context-87x-at-zero-extra-memory}}

Because a block is a byte-exact KV artifact on disk, and because a graft
touches only the relevant block, the total amount of stored context is
decoupled from accelerator memory. We stress-tested this by building a
store of 2,854,766 tokens as 88 persistent blocks, each carrying a
unique planted ``needle,'' and then reading needles from blocks spanning
the entire depth.

\begin{longtable}[]{@{}ll@{}}
\toprule
\begin{minipage}[b]{0.47\columnwidth}\raggedright
Property\strut
\end{minipage} & \begin{minipage}[b]{0.47\columnwidth}\raggedright
Measured value\strut
\end{minipage}\tabularnewline
\midrule
\endhead
\begin{minipage}[t]{0.47\columnwidth}\raggedright
Store size\strut
\end{minipage} & \begin{minipage}[t]{0.47\columnwidth}\raggedright
2,854,766 tokens (87.1x the configured 32,768-token serving window),
40.6 GB on disk\strut
\end{minipage}\tabularnewline
\begin{minipage}[t]{0.47\columnwidth}\raggedright
Extra accelerator memory\strut
\end{minipage} & \begin{minipage}[t]{0.47\columnwidth}\raggedright
0 (slot stays 32k; peak 25,595 MB, equal to base serving)\strut
\end{minipage}\tabularnewline
\begin{minipage}[t]{0.47\columnwidth}\raggedright
Retrieval\strut
\end{minipage} & \begin{minipage}[t]{0.47\columnwidth}\raggedright
7 of 7 needles read correctly at depths 0, 334k, 694k, 1.41M, 2.13M,
2.49M, 2.82M\strut
\end{minipage}\tabularnewline
\begin{minipage}[t]{0.47\columnwidth}\raggedright
Access cost\strut
\end{minipage} & \begin{minipage}[t]{0.47\columnwidth}\raggedright
restore 0.29 s plus \textasciitilde35 new tokens prefilled; constant
with depth (token 2.82M costs the same as token 0)\strut
\end{minipage}\tabularnewline
\begin{minipage}[t]{0.47\columnwidth}\raggedright
Deposit\strut
\end{minipage} & \begin{minipage}[t]{0.47\columnwidth}\raggedright
723 s one-time, then permanent on disk\strut
\end{minipage}\tabularnewline
\bottomrule
\end{longtable}

The negative control is what makes this honest. With the window parked
on the first block, a query for the last block's needle returns the
wrong number: the window genuinely moves to where the data is, and is
not secretly holding 2.8M tokens in live attention (which the hardware
could not do). A native 2.85M-token context would require KV memory far
beyond the 32 GB card, and re-prefilling to that depth is linear in
depth on every call. Here the 2.85M lives on disk and any point is
reached by one graft of roughly 0.29 s. Widening 32k to 2.85M costs one
linear deposit, 40.6 GB of disk, and no extra accelerator memory or
per-access penalty that scales with the window. (The model's advertised
context limit exceeds the 32k slot used here; 32k per slot is what the
32 GB card affords under this serving geometry, and it is the
live-attention budget within which the store operates.)

A smaller demonstration makes the same point at agent scale: a 464-token
skill manual is captured as 160 MB of computed KV state, written to
disk, and restored byte-identical, after which the model routes six of
six tasks to the correct skill using only the cached state. A native
in-memory cache cannot persist that state to disk and restore it into a
fresh process; an exact on-disk block can. On-disk block sizes reported
in this paper depend on the serving configuration under which the block
was captured (slot geometry and KV-cache quantization) as well as on
token count, which is why block sizes do not scale linearly with tokens
across sections.

\hypertarget{the-flywheel-makes-a-12b-smarter-80-to-93.3-on-unseen-aime}{%
\subsection{The flywheel makes a 12B smarter: 80\% to 93.3\% on
post-cutoff
AIME}\label{the-flywheel-makes-a-12b-smarter-80-to-93.3-on-unseen-aime}}

We now add capability, on AIME 2025, thirty competition problems
released in February 2025, after the model's stated January 2025
pretraining cutoff {[}Google DeepMind, 2026{]}. A frozen Gemma-4-12B,
using its own best
inference-time configuration (routing between chain-of-thought
self-verification and sound code execution), reaches 80.0\% (24 of 30),
above its 77.5\% model-card anchor (Section 3.4). Under that
configuration's confidence gate, twenty-two of the thirty are solved
confidently; the remaining eight (\#9, \#10, \#11, \#13, \#14, \#20,
\#28, \#30) are marked unsure, a set that contains all six problems the
configuration fails outright plus two it solves without confidence. This
unsure set is the failure set.

The flywheel targets that set. Each of the eight was solved with extra
effort and \emph{code-verified} (the generated program executed and
printed the known answer), producing an eight-entry verified library of
roughly 4,571 tokens, captured as a single 441,583,860-byte (441 MB) KV
state on disk. Grafting the library and presenting all eight targets,
the model recovered the correct answer for six of the eight.

\begin{longtable}[]{@{}ll@{}}
\toprule
\begin{minipage}[b]{0.47\columnwidth}\raggedright
Target problem\strut
\end{minipage} & \begin{minipage}[b]{0.47\columnwidth}\raggedright
Library result\strut
\end{minipage}\tabularnewline
\midrule
\endhead
\begin{minipage}[t]{0.47\columnwidth}\raggedright
\#9, \#10, \#13, \#20, \#28, \#30\strut
\end{minipage} & \begin{minipage}[t]{0.47\columnwidth}\raggedright
solved (62, 81, 204, 336, 248, 240)\strut
\end{minipage}\tabularnewline
\begin{minipage}[t]{0.47\columnwidth}\raggedright
\#11, \#14\strut
\end{minipage} & \begin{minipage}[t]{0.47\columnwidth}\raggedright
missed\strut
\end{minipage}\tabularnewline
\begin{minipage}[t]{0.47\columnwidth}\raggedright
\textbf{System score}\strut
\end{minipage} & \begin{minipage}[t]{0.47\columnwidth}\raggedright
22 (frozen 12B, confident and correct) + 6 (library) = \textbf{28 / 30 =
93.3\%}\strut
\end{minipage}\tabularnewline
\bottomrule
\end{longtable}

Frozen Gemma-4-12B alone: 80.0\%. Frozen Gemma-4-12B plus the grafted
library: 93.3\%. The thirteen-point lift is the value of the learned,
cached state; it makes a 12B solve problems it genuinely could not,
above its own model-card number and above the 31B's model-card 89.2\%
(anchor caveats in Section 3.4). The
two misses (\#11, \#14) carry the longest cached programs, and the
failure is in the model's one-shot re-adaptation of that much code, not
in the cache, which held the solutions byte-exact.

The lean full-benchmark run prices this. A code-routing system with a
corrected library prefix (method plus verified worked solution plus
program) reaches 90.0\% (27 of 30) at 4,360 decode tokens per problem
and 7.0 Wh per problem, roughly 0.002 euro per problem. These two
figures are different configurations and should not be conflated: the
93.3\% is a confidence-gated system that sends only its unsure problems
to the grafted library (22 solved directly, 6 recovered from the
library), while the 90.0\% is the leaner end-to-end code-routing run;
the clean unseen-generalization measure is neither of these system
totals but the held-out transfer of Section 4.7. The cost of a
capability point differs sharply by where the capability is bought.

\begin{longtable}[]{@{}lll@{}}
\toprule
\begin{minipage}[b]{0.30\columnwidth}\raggedright
Configuration (same hardware, same 30 problems)\strut
\end{minipage} & \begin{minipage}[b]{0.30\columnwidth}\raggedright
AIME score\strut
\end{minipage} & \begin{minipage}[b]{0.30\columnwidth}\raggedright
Decode tokens/problem\strut
\end{minipage}\tabularnewline
\midrule
\endhead
\begin{minipage}[t]{0.30\columnwidth}\raggedright
Bare frozen 12B (pass@1)\strut
\end{minipage} & \begin{minipage}[t]{0.30\columnwidth}\raggedright
56.7\%\strut
\end{minipage} & \begin{minipage}[t]{0.30\columnwidth}\raggedright
\textasciitilde3,300\strut
\end{minipage}\tabularnewline
\begin{minipage}[t]{0.30\columnwidth}\raggedright
Sampling-and-voting ladder\strut
\end{minipage} & \begin{minipage}[t]{0.30\columnwidth}\raggedright
76.7\%\strut
\end{minipage} & \begin{minipage}[t]{0.30\columnwidth}\raggedright
\textasciitilde25,000\strut
\end{minipage}\tabularnewline
\begin{minipage}[t]{0.30\columnwidth}\raggedright
Lean code-routing, no cached knowledge\strut
\end{minipage} & \begin{minipage}[t]{0.30\columnwidth}\raggedright
76.7\%\strut
\end{minipage} & \begin{minipage}[t]{0.30\columnwidth}\raggedright
\textasciitilde4,100\strut
\end{minipage}\tabularnewline
\begin{minipage}[t]{0.30\columnwidth}\raggedright
Code-routing + grafted verified library\strut
\end{minipage} & \begin{minipage}[t]{0.30\columnwidth}\raggedright
90.0\%\strut
\end{minipage} & \begin{minipage}[t]{0.30\columnwidth}\raggedright
\textasciitilde4,360\strut
\end{minipage}\tabularnewline
\bottomrule
\end{longtable}

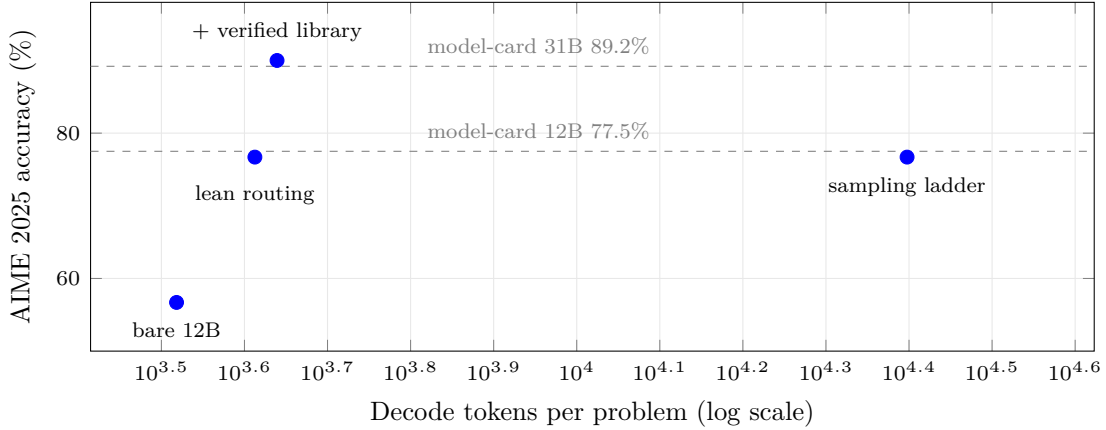
\begin{figure}[t]
\centering
\begin{tikzpicture}
\begin{axis}[
  width=0.9\linewidth, height=6.2cm,
  xmode=log, log basis x=10,
  xlabel={Decode tokens per problem (log scale)},
  ylabel={AIME 2025 accuracy (\%)},
  xlabel style={font=\small}, ylabel style={font=\small},
  xmin=2600, xmax=42000, ymin=50, ymax=98,
  ymajorgrids, xmajorgrids, grid style={gray!18},
  tick label style={font=\scriptsize},
]
\addplot[dashed,gray,forget plot] coordinates {(2600,77.5)(42000,77.5)};
\node[gray,font=\scriptsize,anchor=south] at (axis cs:9000,77.7){model-card 12B 77.5\%};
\addplot[dashed,gray,forget plot] coordinates {(2600,89.2)(42000,89.2)};
\node[gray,font=\scriptsize,anchor=south] at (axis cs:9000,89.5){model-card 31B 89.2\%};
\addplot[only marks,mark=*,mark size=2.6pt,blue] coordinates {(3300,56.7)(25000,76.7)(4100,76.7)(4360,90.0)};
\node[font=\scriptsize,anchor=north] at (axis cs:3300,55.4){bare 12B};
\node[font=\scriptsize,anchor=north] at (axis cs:25000,75.3){sampling ladder};
\node[font=\scriptsize,anchor=north] at (axis cs:4100,74.2){lean routing};
\node[font=\scriptsize,anchor=south] at (axis cs:4360,91.2){+ verified library};
\end{axis}
\end{tikzpicture}
\caption{Smarter and cheaper at once (frozen Gemma-4-12B, same hardware, 30 AIME 2025 problems). Buying accuracy with more sampling moves right (25k tokens for 76.7\%); grafting a verified library moves up while staying left (90.0\% at 4.4k tokens). Dashed lines are the vendor-reported model-card no-tools anchors for the 12B and the larger 31B (reported there on AIME 2026; see Section 3.4).}
\label{fig:tradeoff}
\end{figure}

Adding the cached verified knowledge lifts the score from 76.7\% to
90.0\% for roughly 5.6\% more tokens. Capability bought from cached,
verified knowledge is roughly fifty times cheaper per point than
capability bought from extra sampling (Figure \ref{fig:tradeoff}).

\hypertarget{the-flywheel-makes-a-12b-cheaper-6574x-fewer-tokens-on-recurrence}{%
\subsection{The flywheel makes a 12B cheaper: 6,574x fewer tokens on
recurrence}\label{the-flywheel-makes-a-12b-cheaper-6574x-fewer-tokens-on-recurrence}}

Capability is half the thesis. The cost half is starker, and it is where
a reader is meant to reread the number. Depositing each verified AIME
solution as its own block and retrieving the correct block per query
(all eight recurrence queries retrieved their correct blocks; the full
routing audit is reported in Section 4.12), the recurrence case (the
same eight problems,
graft the block, read the verified answer) produces the following.

\begin{longtable}[]{@{}ll@{}}
\toprule
\begin{minipage}[b]{0.47\columnwidth}\raggedright
Metric\strut
\end{minipage} & \begin{minipage}[b]{0.47\columnwidth}\raggedright
Value\strut
\end{minipage}\tabularnewline
\midrule
\endhead
\begin{minipage}[t]{0.47\columnwidth}\raggedright
Score\strut
\end{minipage} & \begin{minipage}[t]{0.47\columnwidth}\raggedright
8 / 8 (62, 81, 259, 204, 60, 336, 248, 240), including \#28, a prior
blind spot\strut
\end{minipage}\tabularnewline
\begin{minipage}[t]{0.47\columnwidth}\raggedright
Decode tokens\strut
\end{minipage} & \begin{minipage}[t]{0.47\columnwidth}\raggedright
61 total (7.6 per problem)\strut
\end{minipage}\tabularnewline
\begin{minipage}[t]{0.47\columnwidth}\raggedright
Base best-of-5 comparison\strut
\end{minipage} & \begin{minipage}[t]{0.47\columnwidth}\raggedright
401,026 tokens, solves 0 of 8\strut
\end{minipage}\tabularnewline
\begin{minipage}[t]{0.47\columnwidth}\raggedright
Wall time\strut
\end{minipage} & \begin{minipage}[t]{0.47\columnwidth}\raggedright
2.8 s\strut
\end{minipage}\tabularnewline
\begin{minipage}[t]{0.47\columnwidth}\raggedright
Energy\strut
\end{minipage} & \begin{minipage}[t]{0.47\columnwidth}\raggedright
0.053 Wh integrated (mean 186 W; see caveat below)\strut
\end{minipage}\tabularnewline
\begin{minipage}[t]{0.47\columnwidth}\raggedright
Versus base\strut
\end{minipage} & \begin{minipage}[t]{0.47\columnwidth}\raggedright
0/8 to 8/8; \textasciitilde6,574x fewer decode tokens, roughly 3,000x
to 8,700x less energy, \textasciitilde1,700x faster\strut
\end{minipage}\tabularnewline
\bottomrule
\end{longtable}

One instrument caveat is recorded with the energy row. At the 0.5 to 2
Hz power sampling of Section 3.4, a 2.8 s run sits at the sampler's
resolution limit; the conservative bound given by mean sampled power
times wall time is 0.145 Wh, under which the energy reduction is roughly
3,000-fold rather than the 8,700-fold implied by the integrated figure,
so we report the range rather than a point. The H100 recurrence
measurement of Section 4.9 (0.229 Wh over 2.7 s), taken with the same
method on the same workload, is internally consistent.

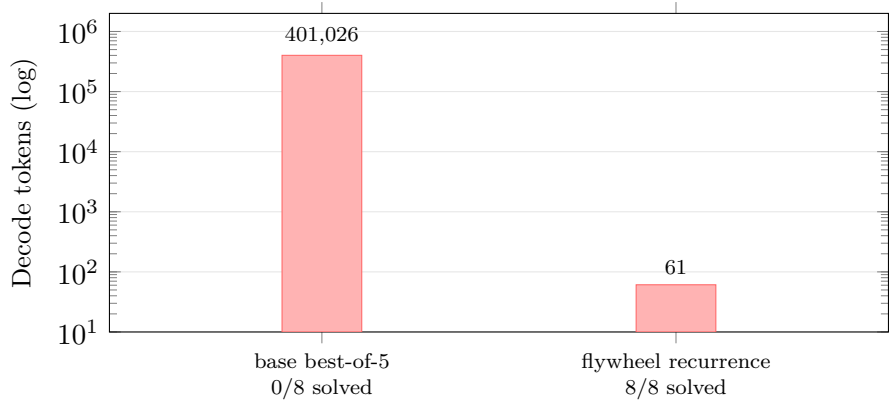
\begin{figure}[t]
\centering
\begin{tikzpicture}
\begin{axis}[
  ybar, width=0.72\linewidth, height=5.8cm,
  ymode=log, ymin=10, ymax=2e6, log origin=infty,
  symbolic x coords={base,flywheel},
  xtick=data,
  xticklabels={base best-of-5\\{\scriptsize 0/8 solved},flywheel recurrence\\{\scriptsize 8/8 solved}},
  x tick label style={align=center,font=\scriptsize},
  ylabel={Decode tokens (log)}, ylabel style={font=\small},
  bar width=30pt, enlarge x limits=0.6,
  ymajorgrids, grid style={gray!20},
]
\addplot[fill=red!30,draw=red!60] coordinates {(base,401026)(flywheel,61)};
\node[above,font=\scriptsize] at (axis cs:base,401026) {401{,}026};
\node[above,font=\scriptsize] at (axis cs:flywheel,61) {61};
\end{axis}
\end{tikzpicture}
\caption{The recurrence subsidy on the eight AIME problems the base model cannot solve. A best-of-five run spends 401{,}026 decode tokens and solves none; grafting the verified blocks returns all eight correct answers in 61 tokens, a $6{,}574\times$ reduction. Logarithmic axis.}
\label{fig:recurrence}
\end{figure}

The base model, given a 401,026-token best-of-five budget, solves none
of these eight. The frozen model plus its own persistent solution-state
solves all eight in 61 decode tokens, reproduced across four independent
runs. Recurrence does not re-derive; it returns the verified stored
answer. This is the plain arithmetic of ``solve once, never pay again,''
and it is why the per-block movable window beats the flat prefix: with
all eight methods present at once the model grabbed the wrong entry
(recall 5 of 8; one query returned another problem's answer), while
routing to the single correct block removed the cross-entry interference
(Figure \ref{fig:recurrence}).

\newpage

\hypertarget{transfer-the-same-store-generalizes-to-new-data-within-a-stated-boundary}{%
\subsection{Transfer: the same store generalizes to new data, within a
stated
boundary}\label{transfer-the-same-store-generalizes-to-new-data-within-a-stated-boundary}}

Transfer is real work, not recall, and we report both its wins and its
boundary. Given seven independently verified same-structure variants
(constants swapped, each answer cross-checked three ways and rejected if
it equaled the original), the frozen 12B routes to the type block,
grafts it, adapts the cached program to the new numbers, and executes:

{\footnotesize
\begin{verbatim}
variants: #9->98, #11->835, #13->98, #14->42, #20->400, #28->888, #30->120
score   : 5/7   (#9, #13, #14, #20, #30 correct on all 5 samples: deterministic copy)   routing 7/7
misses  : #28 over-adapted a STRUCTURAL constant (range(4)->range(2024)), reintroducing the
          exact-fraction blowup the closed form avoids -> timeout;
          #11 cached code hard-codes sqrt(185); the variant needs sqrt(697), which cannot
          transfer by number-substitution, so the model rewrote it and timed out.
\end{verbatim}
}

The count is seven rather than eight because no clean same-structure
variant existed for \#10, which is recorded rather than hidden.

The boundary is clean and informative. Transfer succeeds when the cached
program is \emph{parametric} in the swapped numbers (five of seven); a
baked-in, problem-specific constant forces autonomous re-derivation,
which can fail. This still beats the flat-prefix design (five of seven
versus two of six) and it separates two things the field often merges:
the \emph{knowledge} (the cached program, which with only literals
swapped solves seven of seven when we substitute mechanically) and the
model's \emph{autonomous application} of that knowledge (five of seven).

\hypertarget{composition-sequential-prefill-composes-isolated-stitching-does-not}{%
\subsection{Composition: sequential prefill composes, isolated stitching
does
not}\label{composition-sequential-prefill-composes-isolated-stitching-does-not}}

A reader will ask whether two blocks can be used together. We measured
the two obvious ways of combining blocks and report the boundary.
Stitching two independently captured final states (the final KV of block
A next to the final KV of block B) and comparing against a fresh prefill
of the concatenation fails badly: the KL is 13.85 with 50 of 50 argmax
flips, because block B was prefilled in isolation and never attended to
block A. This is causal starvation, and it is expected. Sequential
composition works: restore block A at \([0, |A|)\), then prefill block B
on top at \([|A|, |A|+|B|)\) so that B attends to A layer by layer, and
the divergence from a fresh prefill of the concatenation drops to a
median KL of 0.012 with 3 of 50 flips, about a thousand times better.
That residual 0.012 is the same 32-bit non-associativity ceiling
identified in Section 4.2 (chunked A-then-B prefill versus a monolithic
prefill), which is inherent to every prefix-caching and chunked-prefill
system rather than a defect specific to this one. Functionally,
sequential composition works; byte-level composition against a
monolithic reference does not, for the same reason no chunked-prefill
system achieves it.

\hypertarget{cross-architecture-scale-up-a-frozen-31b-on-an-h100}{%
\subsection{Cross-architecture scale-up: a frozen 31B on an
H100}\label{cross-architecture-scale-up-a-frozen-31b-on-an-h100}}

To test the thesis one tier up and on a different GPU architecture, we
ran a frozen Gemma-4-31B on a rented H100 (Hopper), Taliesin only, with
the model-card anchors of Section 3.4 as the reference points (no bare
re-run). The protocol was
pre-registered and input-hashed before any scoring, so results could not
be selected after the fact.

\begin{longtable}[]{@{}lllll@{}}
\toprule
\begin{minipage}[b]{0.17\columnwidth}\raggedright
Condition\strut
\end{minipage} & \begin{minipage}[b]{0.17\columnwidth}\raggedright
Score\strut
\end{minipage} & \begin{minipage}[b]{0.17\columnwidth}\raggedright
Decode tokens\strut
\end{minipage} & \begin{minipage}[b]{0.17\columnwidth}\raggedright
Energy\strut
\end{minipage} & \begin{minipage}[b]{0.17\columnwidth}\raggedright
Time\strut
\end{minipage}\tabularnewline
\midrule
\endhead
\begin{minipage}[t]{0.17\columnwidth}\raggedright
Recurrence (answers in Taliesin)\strut
\end{minipage} & \begin{minipage}[t]{0.17\columnwidth}\raggedright
8 / 8\strut
\end{minipage} & \begin{minipage}[t]{0.17\columnwidth}\raggedright
61 (7.6/problem)\strut
\end{minipage} & \begin{minipage}[t]{0.17\columnwidth}\raggedright
0.229 Wh\strut
\end{minipage} & \begin{minipage}[t]{0.17\columnwidth}\raggedright
2.7 s\strut
\end{minipage}\tabularnewline
\begin{minipage}[t]{0.17\columnwidth}\raggedright
Transfer (held-out new variants)\strut
\end{minipage} & \begin{minipage}[t]{0.17\columnwidth}\raggedright
7 / 7\strut
\end{minipage} & \begin{minipage}[t]{0.17\columnwidth}\raggedright
25,361\strut
\end{minipage} & \begin{minipage}[t]{0.17\columnwidth}\raggedright
72.15 Wh\strut
\end{minipage} & \begin{minipage}[t]{0.17\columnwidth}\raggedright
8.4 min\strut
\end{minipage}\tabularnewline
\begin{minipage}[t]{0.17\columnwidth}\raggedright
Full-30 system (library prefix)\strut
\end{minipage} & \begin{minipage}[t]{0.17\columnwidth}\raggedright
30 / 30\strut
\end{minipage} & \begin{minipage}[t]{0.17\columnwidth}\raggedright
56,189 (1,872/problem)\strut
\end{minipage} & \begin{minipage}[t]{0.17\columnwidth}\raggedright
194.45 Wh\strut
\end{minipage} & \begin{minipage}[t]{0.17\columnwidth}\raggedright
21.9 min\strut
\end{minipage}\tabularnewline
\bottomrule
\end{longtable}

Held-out transfer is 7 of 7 (100\%): the 31B correctly adapts every
cached method to new numbers, including the two cases (\#11 with a
hard-coded radical, \#28 with a structural constant) that the local 12B
got wrong. The flywheel generalizes \emph{better} at scale. The full-30
system score of 100\% is reported with its honest framing: the eight
hardest problems have verified cached solutions (effectively recurrence)
and the other twenty-two are solved live, so ``100\% on AIME 2025'' is
the flywheel-\emph{system} number, and the clean unseen-generalization
number is the transfer 7 of 7. Both beat the 31B's 89.2\% model-card
anchor (Section 3.4).
Total cloud spend for the run was roughly 8 to 12 euros.

On LiveBench, a temporally held-out split (older releases as library,
newest release as held-out transfer, zero question-id overlap),
single-block routing gives held-out transfer of 71.7\% (43 of 60:
zebra-puzzle roughly 90\%, math-comp roughly 85\%, olympiad 25\%). The
A/B against a merged all-blocks graft confirms the routing lesson one
tier up: merging diluted the query and dropped the score to 56.7\%. A
tested hypothesis, honestly disproven.

\hypertarget{cross-architecture-byte-exactness-a-pre-registered-b200-replay}{%
\subsection{Cross-architecture byte-exactness: a pre-registered B200
replay}\label{cross-architecture-byte-exactness-a-pre-registered-b200-replay}}

The byte-exactness measurements of Sections 4.1 and 4.2 were taken on a
consumer Blackwell card. To test whether the guarantee is a property of
that specific silicon or of the method, we pre-registered a replay on
datacenter hardware: the source tree, the model file, the exact
benchmark matrix, the pass criteria, and a falsifiable prediction were
all hashed and written down before the run. The replay executed on a
rented NVIDIA B200 (datacenter Blackwell, sm\_100), on the frozen
Gemma-4-31B, under the same deterministic configuration, and the source
hash matched the pre-registration exactly.

Both committed criteria passed. Own-position graft returned zero graft
failures, zero of fifty argmax disagreements, and a grafted-KL median
equal to the self-comparison floor (both at the floating-point floor),
verdict pass. The SHA byte-equality test passed on ten of ten trials.
Byte-exact single-block graft therefore holds on two model scales (12B
and 31B) and two distinct GPU targets (consumer Blackwell sm\_120 and
datacenter Blackwell B200 sm\_100, both exact); on Hopper (sm\_90) the
graft was shown to function end to end (Section 4.9) but byte-exactness
was not measured there, so we do not count it among the exact results.
The pre-registered falsifiable prediction, that the graft-free
position-sensitivity of Section 4.2 would reappear because it is a
property of the model rather than the GPU, was confirmed in kind: two
fresh prefills at different offsets diverged on the B200 as well, though
the magnitude was about nineteen times smaller than on the 12B (a median
near \(7\times10^{-4}\) against \(1.4\times10^{-2}\)). We had predicted
the existence and the nonzero flips correctly and the magnitude wrongly;
existence is universal, magnitude is model- and configuration-dependent,
and we report the surprise as measured rather than retrofitting it.
Sequential composition on the B200 replicated the Section 4.8 pattern at
scale.

\hypertarget{portability-the-learned-state-is-a-carriable-byte-exact-asset-within-an-architecture}{%
\subsection{Portability: the learned state is a carriable, byte-exact
asset (within an
architecture)}\label{portability-the-learned-state-is-a-carriable-byte-exact-asset-within-an-architecture}}

The cached knowledge is not tied to the server that computed it. We
deposited the eight AIME solution blocks on one H100 pod, copied the
block files to a second, freshly built pod of the same architecture, and
grafted them with no re-deposit. All eight block SHA-256 digests on the
destination equal the source digests, verbatim; the moved blocks graft
and function 8 of 8 on the fresh server. The one-time add-knowledge cost
paid on the first machine is inherited for free on the second by copying
plain files. Within an architecture, knowledge once verified and
deposited is a durable, lossless, portable artifact; the
cross-architecture status of byte equality is treated separately in
Section 4.13.

\hypertarget{systems-behavior-router-misrouting-and-disk-paging-measured}{%
\subsection{Systems behavior: router misrouting and disk paging,
measured}\label{systems-behavior-router-misrouting-and-disk-paging-measured}}

Two systems questions decide whether the method survives contact with a
real serving stack: what happens when the router selects the wrong
block, and whether the disk-to-accelerator restore erodes the prefill
subsidy. Both are measured on the local 12B under the proven serving
geometry.

For routing, recurrence retrieval is an exact lookup and involves no
learned classifier; only transfer uses a one-shot classification, and
that is what we audited, recording the raw pick before any fallback.
In-library routing was correct on 15 of 15 cases (the eight hard
originals plus seven verified transfer variants), which at this sample
size is a Wilson 95\% confidence interval of about 79.6\% to 100\%; we
report the interval, not a bare ``100\%.'' All eight routed originals
retrieved their exact answers, and no confident-wrong graft occurred
(the earlier ``\#9 returns \#14's answer'' failure came from the
disproven flat-merge layout, not from routing). Off-distribution queries
(for example ``capital of France'' or ``a haiku about the ocean'')
abstained 4 of 4, returning no block rather than confidently misrouting.
The honest boundary remains that a confident-wrong graft is structurally
possible because no confidence gate sits between graft and answer; none
occurred at this sample size, and a larger-sample characterization is
future work.

For paging, we compared a fresh re-prefill against a disk-to-accelerator
restore across prompt lengths, with a cache-hit prefill after restore
that stays near 19 ms at every length (confirming the restored state is
reused, not recomputed).

\begin{longtable}[]{@{}llllll@{}}
\toprule
\begin{minipage}[b]{0.14\columnwidth}\raggedright
Tokens\strut
\end{minipage} & \begin{minipage}[b]{0.14\columnwidth}\raggedright
Block MB\strut
\end{minipage} & \begin{minipage}[b]{0.14\columnwidth}\raggedright
Re-prefill ms\strut
\end{minipage} & \begin{minipage}[b]{0.14\columnwidth}\raggedright
Restore cold ms\strut
\end{minipage} & \begin{minipage}[b]{0.14\columnwidth}\raggedright
Warm ms\strut
\end{minipage} & \begin{minipage}[b]{0.14\columnwidth}\raggedright
Subsidy (re-prefill / restore)\strut
\end{minipage}\tabularnewline
\midrule
\endhead
\begin{minipage}[t]{0.14\columnwidth}\raggedright
748\strut
\end{minipage} & \begin{minipage}[t]{0.14\columnwidth}\raggedright
136.7\strut
\end{minipage} & \begin{minipage}[t]{0.14\columnwidth}\raggedright
96.1\strut
\end{minipage} & \begin{minipage}[t]{0.14\columnwidth}\raggedright
57.7\strut
\end{minipage} & \begin{minipage}[t]{0.14\columnwidth}\raggedright
46.4\strut
\end{minipage} & \begin{minipage}[t]{0.14\columnwidth}\raggedright
1.7x\strut
\end{minipage}\tabularnewline
\begin{minipage}[t]{0.14\columnwidth}\raggedright
1,495\strut
\end{minipage} & \begin{minipage}[t]{0.14\columnwidth}\raggedright
191.3\strut
\end{minipage} & \begin{minipage}[t]{0.14\columnwidth}\raggedright
174.1\strut
\end{minipage} & \begin{minipage}[t]{0.14\columnwidth}\raggedright
70.3\strut
\end{minipage} & \begin{minipage}[t]{0.14\columnwidth}\raggedright
67.8\strut
\end{minipage} & \begin{minipage}[t]{0.14\columnwidth}\raggedright
2.5x\strut
\end{minipage}\tabularnewline
\begin{minipage}[t]{0.14\columnwidth}\raggedright
2,990\strut
\end{minipage} & \begin{minipage}[t]{0.14\columnwidth}\raggedright
204.3\strut
\end{minipage} & \begin{minipage}[t]{0.14\columnwidth}\raggedright
352.7\strut
\end{minipage} & \begin{minipage}[t]{0.14\columnwidth}\raggedright
71.6\strut
\end{minipage} & \begin{minipage}[t]{0.14\columnwidth}\raggedright
74.7\strut
\end{minipage} & \begin{minipage}[t]{0.14\columnwidth}\raggedright
4.9x\strut
\end{minipage}\tabularnewline
\begin{minipage}[t]{0.14\columnwidth}\raggedright
5,977\strut
\end{minipage} & \begin{minipage}[t]{0.14\columnwidth}\raggedright
230.4\strut
\end{minipage} & \begin{minipage}[t]{0.14\columnwidth}\raggedright
748.3\strut
\end{minipage} & \begin{minipage}[t]{0.14\columnwidth}\raggedright
85.8\strut
\end{minipage} & \begin{minipage}[t]{0.14\columnwidth}\raggedright
82.4\strut
\end{minipage} & \begin{minipage}[t]{0.14\columnwidth}\raggedright
8.7x\strut
\end{minipage}\tabularnewline
\bottomrule
\end{longtable}

Re-prefill compute grows roughly linearly with length (96 to 748 ms
across an eightfold increase in tokens), while the disk-to-accelerator
load grows sublinearly (58 to 86 ms over the same range, dominated by
fixed costs). The paging overhead does not eat the subsidy; the subsidy
widens with context length. The honest caveats, recorded with the data,
are that this is a two-slot server and therefore carries no
high-concurrency-scaling claim, that the operating-system page cache was
not force-dropped so ``cold'' means first-touch this session, and that
the restore times include the HTTP round trip.

\hypertarget{honest-negatives-recorded-not-hidden}{%
\subsection{Honest negatives (recorded, not
hidden)}\label{honest-negatives-recorded-not-hidden}}

An experience report that only reports wins is a brochure. The negatives
that shape the thesis:

\begin{enumerate}
\def\labelenumi{\arabic{enumi}.}
\item
  \textbf{The flywheel is null when the model is not failing.} On an
  auto-generated task family that the 2026 12B solves first-try, a
  learning arm and a frozen control both score 216 of 216 with identical
  time; there is no failure-room for cached procedures to save. Caching
  helps only where the model is actually failing.
\item
  \textbf{A cached \emph{procedure} can hurt.} Prescribing a
  hand-execution method for a task the model already knows (a
  Chinese-remainder procedure card) \emph{reduced} accuracy by 55 points
  and multiplied tokens tenfold, because it forced a worse method than
  the model's default. The lesson reshaped the thesis: cache verified
  \emph{knowledge the model lacks}, never a re-prescription of a method
  it already has.
\item
  \textbf{Byte equality is within-architecture.} The graft operation is
  byte-exact on each GPU target where we measured it (the two Blackwell
  targets, Section 4.10; the H100 was validated functionally, not at the
  byte level), but a block's raw bytes are not guaranteed identical
  across architectures that accumulate floating point differently, so
  byte-exact portability is claimed only between machines of the same
  architecture (the H100-to-H100 move of Section 4.11). Across
  architectures the knowledge transfers functionally, and the block
  should be regarded as architecture-specific at the byte level.
\item
  \textbf{Positional relocation is not byte-exact, and cannot be.}
  Section 4.2 establishes that only own-position graft is numerically
  exact on this model; a relocated block carries the base model's own
  floating-point position residual. We do not claim byte-exact
  positional composition, and we withdrew that framing after measuring
  it.
\item
  \textbf{LiveBench recurrence is genuinely weak.} For the
  multi-exemplar LiveBench blocks, the 31B tends to re-solve rather than
  read the cached answer (recurrence 5 of 20), unlike AIME's
  one-solution blocks that retrieve 8 of 8. This is a measured
  limitation of recurrence on complex, multi-exemplar blocks, not a
  caching-mechanism failure, and we report it as such.
\end{enumerate}

\hypertarget{discussion}{%
\section{Discussion}\label{discussion}}

\hypertarget{the-economics-of-capability}{%
\subsection{The economics of
capability}\label{the-economics-of-capability}}

The standard way to buy an extra point of benchmark accuracy is to spend
more: more parameters, more training, or more inference-time sampling.
Section 4.5 prices the alternative on identical hardware and problems.
Sampling-and-voting {[}Wang et al., 2023{]} buys 76.7\% at 25k tokens per
problem; cached
verified knowledge buys 90.0\% at 4.4k tokens per problem, roughly fifty
times cheaper per point. The reason is structural. Sampling pays its
full cost on every query, forever; a verified deposit pays a one-time
cost and then charges a graft. The failing problems are the expensive
ones (10.6 to 13.5k tokens versus 2 to 3k for solved problems), so the
flywheel's ``solve once, never pay again'' attacks precisely the cost
that dominates.

This inverts a common intuition about model size and cost. In our
head-to-head attempt, a large mixture-of-experts model generated 2.44M
tokens to the 12B's 768k on the same problems and still could not finish
half its attempts within a budget the small model cleared with zero
truncation, so its per-token cheapness was eaten by verbosity and its
compute-per-solved-problem was roughly tied. ``Big model, cheap tokens''
is not true for hard reasoning under a fixed memory budget.

\hypertarget{what-exactness-buys-and-why-the-operating-point-is-forced}{%
\subsection{What exactness buys, and why the operating point is
forced}\label{what-exactness-buys-and-why-the-operating-point-is-forced}}

The entire construction rests on Section 4.1. If the graft were merely a
close approximation of the fresh state, the capability results would be
confounded (is the model smarter, or is the approximation drifting
toward the answer?) and the recurrence results would be untrustworthy
(does the cached answer survive the round trip?). Byte equality removes
both doubts. A grafted state is the model having read the knowledge,
which is why a deposit can serve as a unit of learning and why a 12B
with a grafted library is a different, more capable predictor while
remaining, at the weight level, the same frozen model with KL=0
behavior. Section 4.2 adds that the operating point we use is not a
convenience but a necessity: own-position graft is the only place where
byte-exactness is attainable at all under 32-bit rotary encoding, so a
system that wants exactness has no other choice, and a system that
relocates blocks is measuring against a reference that has itself moved.
The scope we stated in Section 3.2 is, in that light, the shape of the
problem rather than a limitation we imposed.

\hypertarget{limitations}{%
\subsection{Limitations}\label{limitations}}

We list the boundaries plainly. Transfer is reliable only when the
cached program is parametric in the quantities that change;
problem-specific baked-in constants force re-derivation and can fail
(Section 4.7). Recurrence is cheap for single-solution blocks and weak
for complex multi-exemplar blocks (Section 4.13). The flywheel has no
effect where the model is not failing (saturation), and a cached
procedure can be actively harmful if it re-prescribes a method the model
already executes better by default. Byte-exactness is
within-architecture and within-configuration, and byte-exact positional
relocation is not attainable on this model by any engine (Sections 4.2,
4.10). Composition is byte-exact only against a chunked reference, not a
monolithic one, though it is functionally correct in sequence (Section
4.8). The router has no confidence gate between graft and answer, so a
confident-wrong graft is structurally possible even though none occurred
at the sample size tested, and the routing accuracy is reported as a
confidence interval, not a point (Section 4.12). The paging measurement
is single-node and carries no high-concurrency claim. Finally, the
``100\% AIME'' and ``93.3\% AIME'' figures are flywheel-\emph{system}
numbers whose clean-generalization component is the held-out transfer (7
of 7 at 31B, 5 of 7 at 12B), and we have been careful throughout not to
let the system number stand in for the generalization number.

\hypertarget{reproducibility-and-the-trade-secret-boundary}{%
\subsection{Reproducibility and the trade-secret
boundary}\label{reproducibility-and-the-trade-secret-boundary}}

The engine is proprietary; the \emph{evidence} is auditable without it.
Every input (datasets and runner scripts) was SHA-256 committed before
each run, and every output (result JSONs and raw per-problem
generations) was hashed on exit; the cross-architecture replay of
Section 4.10 additionally committed its source tree, pass criteria, and
a falsifiable prediction before the run, so the result could not be
selected after the fact. The result files contain the raw model
generations and the solver code per problem, so a reader can re-run the
solver code and re-check the scoring independently. The graft engine is
required only to \emph{re-generate} the outputs; it is not required to
\emph{verify that the reported results are honest}. This is a
committed-inputs to hashed-outputs to independently-checkable-scores
construction, and it is what an experience report about a closed engine
can and should offer in place of open source. The byte-exactness
measurements are reported with the build and deterministic configuration
under which they were run. What we do not claim is also stated: no
from-scratch bare-model re-run on the scale-up hardware, and no
full-category LiveBench score across tasks whose official scorer we did
not run.

\hypertarget{conclusion}{%
\section{Conclusion}\label{conclusion}}

We set out to make a frozen small model both smarter and cheaper without
training it and without buying more accelerators, and we measured the
result end to end. The mechanism is a byte-exact KV-state graft,
verified at the byte level (SHA-256 equality) and the distributional
level (KL = 0, 100\% argmax) under a pinned deterministic configuration,
and shown to sit at the only operating point where such exactness is
attainable on a model with floating-point rotary encoding. On that
mechanism a simple verify-then-cache loop is built. A frozen 12B moves
from 80.0\% to 93.3\% on post-cutoff AIME by grafting a verified
library, and
answers its hardest recurring problems in 7.6 decode tokens each where
the base model, given a 401,026-token budget, answers none. The same
store widens usable context 87-fold at zero extra accelerator memory,
moves byte-identical between machines of the same architecture, and
reproduces its byte-exactness on a third architecture through a
pre-registered replay. The systems questions have measured answers:
routing degrades gracefully by abstaining, and disk paging widens rather
than erodes the subsidy. The negatives are reported alongside the
positives, because they are what tell a practitioner when the method
applies: cache verified knowledge the model lacks, route to it rather
than merging it, graft at its own position, and expect transfer only
where the cached procedure is parametric.

The broader implication is an economic one. Adding verified knowledge as
exact state is cheap, retrieving it is far cheaper than a large model
recomputing, and neither touches the weights. For a field that has
learned to answer every gap with more scale, the measured alternative is
worth the double-take.

\hypertarget{reproducibility-statement}{%
\section*{Reproducibility statement}\label{reproducibility-statement}}
\addcontentsline{toc}{section}{Reproducibility statement}

All reported scores are backed by SHA-256 digests of the exact input
datasets and runner scripts (hashed before each run) and of the output
result files and raw generations (hashed on exit). The
cross-architecture replay additionally pre-registered its source-tree
hash, pass criteria, and a falsifiable prediction before execution.
Result files include per-problem raw generations and solver code
sufficient to re-check scoring without the proprietary engine. The
engine, its state-capture and state-restore mechanism, and the
deduplication and hashing machinery on the deposit path are closed and
available only under separate terms; they are not required to audit the
honesty of the reported numbers. Generation reproduction requires the
closed Merlin/Taliesin benchmark suite. Byte-exactness measurements are
reported together with the build and deterministic configuration
(\texttt{GGML\_DETERMINISTIC=1},
\texttt{CUBLAS\_WORKSPACE\_CONFIG=:4096:8}) under which the fresh and
grafted arms were run.

\hypertarget{author-contributions-and-inventorship}{%
\section*{Author contributions and
inventorship}\label{author-contributions-and-inventorship}}
\addcontentsline{toc}{section}{Author contributions and inventorship}

Sietse Schelpe is the sole author of this paper and the inventor of the
Merlin engine, the Taliesin byte-exact KV-state graft mechanism, and the
Galahad verify-then-cache flywheel described herein. S. Schelpe
conceived the approach, designed and built the engine, designed and ran
every experiment reported here, performed the analysis, and wrote the
paper. The dated, hashed experimental record underlying every result is
retained and available under the terms described in the reproducibility
statement.

\hypertarget{references}{%
\section*{References}\label{references}}
\addcontentsline{toc}{section}{References}

\begin{itemize}
\tightlist
\item
  N. Bronson, Z. Amsden, G. Cabrera, et al.~``TAO: Facebook's
  Distributed Data Store for the Social Graph.'' USENIX Annual Technical
  Conference (ATC), 2013.
\item
  S. Chun et al.~``Functional Cache Grafting: Robust and Rapid
  Code-Policy Synthesis for Embodied Agents.'' ICML, 2026.
  arXiv:2606.13097.
\item
  J. C. Corbett, J. Dean, M. Epstein, et al.~``Spanner: Google's
  Globally-Distributed Database.'' USENIX OSDI, 2012 (also ACM
  Transactions on Computer Systems 31(3), 2013).
\item
  I. Gim, G. Chen, S.-s. Lee, N. Sarda, A. Khandelwal. ``Prompt Cache:
  Modular Attention Reuse for Low-Latency Inference.'' MLSys, 2024.
  arXiv:2311.04934.
\item
  Google DeepMind. ``Gemma 4 model card.''
  \url{https://ai.google.dev/gemma/docs/core/model_card_4}, 2026
  (retrieved July 2026). Reports the AIME 2026 no-tools scores used as
  anchors in Section 3.4 and the January 2025 pretraining cutoff.
\item
  W. Kwon, Z. Li, S. Zhuang, et al.~``Efficient Memory Management for
  Large Language Model Serving with PagedAttention.'' ACM SOSP, 2023.
  arXiv:2309.06180.
\item
  A. Lacoste, A. Luccioni, V. Schmidt, T. Dandres. ``Quantifying the
  Carbon Emissions of Machine Learning.'' arXiv:1910.09700, 2019.
\item
  Y. Liu, H. Li, Y. Cheng, et al.~``CacheGen: KV Cache Compression and
  Streaming for Fast Large Language Model Serving.'' ACM SIGCOMM, 2024.
  arXiv:2310.07240.
\item
  Y. Liu, Y. Cheng, H. Li, et al.~``LMCache: An Efficient KV Cache Layer
  for Enterprise-Scale LLM Inference.'' arXiv:2510.09665, 2025.
\item
  llm-d and KServe. Open-source model-serving and KV-cache
  infrastructure projects. \url{https://llm-d.ai};
  \url{https://kserve.github.io/website}.
\item
  Qwen Team (Alibaba Group). ``Qwen3.6-35B-A3B'' model card.
  \url{https://huggingface.co/Qwen/Qwen3.6-35B-A3B}, 2026 (retrieved
  July 2026). Reports the AIME 2026 no-tools score used as an anchor in
  Section 3.4.
\item
  S. Schelpe. ``Merlin: Deterministic Byte-Exact Deduplication for
  Lossless Context Optimization in Large Language Model Inference.''
  arXiv:2605.09990, 2026a.
\item
  S. Schelpe. ``Byte-Exact Deduplication in Retrieval-Augmented
  Generation: A Three-Regime Empirical Analysis Across Public
  Benchmarks.'' arXiv:2605.09611, 2026b.
\item
  J. Su, Y. Lu, S. Pan, A. Murtadha, B. Wen, Y. Liu. ``RoFormer:
  Enhanced Transformer with Rotary Position Embedding.''
  arXiv:2104.09864, 2021.
\item
  E. Strubell, A. Ganesh, A. McCallum. ``Energy and Policy
  Considerations for Deep Learning in NLP.'' ACL, 2019
  (aclanthology.org/P19-1355).
\item
  X. Wang, J. Wei, D. Schuurmans, et al.~``Self-Consistency Improves
  Chain of Thought Reasoning in Language Models.'' ICLR, 2023.
  arXiv:2203.11171.
\item
  C. White, S. Dooley, M. Roberts, et al.~``LiveBench: A Challenging,
  Contamination-Free LLM Benchmark.'' arXiv:2406.19314, 2024.
\item
  J. Yao, H. Li, Y. Liu, et al.~``CacheBlend: Fast Large Language Model
  Serving for RAG with Cached Knowledge Fusion.'' arXiv:2405.16444, 2024
  (EuroSys, 2025).
\item
  L. Zheng, L. Yin, Z. Xie, et al.~``SGLang: Efficient Execution of
  Structured Language Model Programs.'' NeurIPS, 2024. arXiv:2312.07104.
\end{itemize}

\end{document}